\documentclass[sigconf,screen, nonacm,review=false,timestamp=false]{acmart}
\usepackage{svg}
\usepackage{multirow}

\AtBeginDocument{%
  }

\setcopyright{acmcopyright}
\copyrightyear{2018}
\acmYear{2018}
\acmDOI{XXXXXXX.XXXXXXX}

\acmConference[Conference acronym 'XX]{Make sure to enter the correct
  conference title from your rights confirmation emai}{June 03--05,
  2018}{Woodstock, NY}
\acmPrice{15.00}
\acmISBN{978-1-4503-XXXX-X/18/06}

\begin{document}

\title{Contrastive News and Social Media Linking using BERT for Articles and Tweets across Dual Platforms}

\author{Jan Piotrowski}
\email{jf.piotrowsk@student.uw.edu.pl}
\affiliation{%
  \institution{University of Warsaw,}
  \institution{MIM Solutions}
  \country{Poland}
}

\author{Marek Wachnicki}
\email{marek.wachnicki@mim.ai}
\affiliation{%
  \institution{MIM Solutions}
  \country{Poland}
}

\author{Mateusz Perlik}
\email{ms.perlik@student.uw.edu.pl}
\affiliation{%
  \institution{University of Warsaw,}
  \institution{MIM Solutions}
  \country{Poland}
}

\author{Jakub Podolak}
\email{jakub.podolak@student.uva.nl}
\affiliation{%
  \institution{University of Amsterdam}
  \country{The Netherlands}
}

\author{Grzegorz Rucki}
\email{grzegorz.rucki@mim.ai}
\affiliation{%
  \institution{MIM Solutions}
  \country{Poland}
}

\author{Michał Brzozowski}
\email{michal.brzozowski@mim.ai}
\affiliation{%
  \institution{MIM Solutions}
  \country{Poland}
}

\author{Paweł Olejnik}
\email{pw.olejnik@student.uw.edu.pl}
\affiliation{%
  \institution{University of Warsaw}
  \country{Poland}
}

\author{Julian Kozłowski}
\email{jm.kozlowski@student.uw.edu.pl}
\affiliation{%
  \institution{University of Warsaw}
  \country{Poland}
}

\author{Tomasz Nocoń}
\email{t.nocon@student.uw.edu.pl}
\affiliation{%
  \institution{University of Warsaw}
  \country{Poland}
}

\author{Jakub Kozieł}
\email{jakub.koziel.stud@pw.edu.pl}
\affiliation{%
  \institution{Warsaw University of Technology}
  \country{Poland}
}

\author{Stanisław Giziński}
\email{s.gizinski@student.uw.edu.pl}
\affiliation{%
  \institution{University of Warsaw}
  \country{Poland}
}

\author{Piotr Sankowski}
\email{sank@mimuw.edu.pl}
\affiliation{%
  \institution{University of Warsaw, IDEAS NCBR,}
  \institution{MIM Solutions}
\country{Poland}
}

\renewcommand{\shortauthors}{Piotrowski et al.}

\begin{abstract}
X (formerly Twitter) has evolved into a contemporary agora, offering a platform for individuals to express opinions and viewpoints on current events. The majority of the topics discussed on Twitter are directly related to ongoing events, making it an important source for monitoring public discourse. However, linking tweets to specific news presents a significant challenge due to their concise and informal nature. Previous approaches, including topic models, graph-based models, and supervised classifiers, have fallen short in effectively capturing the unique characteristics of tweets and articles.

Inspired by the success of the CLIP model in computer vision, which employs contrastive learning to model similarities between images and captions, this paper introduces a contrastive learning approach for training a representation space where linked articles and tweets exhibit proximity. We present our contrastive learning approach, CATBERT (Contrastive Articles Tweets BERT), leveraging pre-trained BERT models. The model is trained and tested on a dataset containing manually labeled English and Polish tweets and articles related to the Russian-Ukrainian war. 
We evaluate CATBERT's performance against traditional approaches like LDA, and the novel method based on OpenAI embeddings, which has not been previously applied to this task. Our findings indicate that CATBERT demonstrates superior performance in associating tweets with relevant news articles. Furthermore, we demonstrate the performance of the models when applied to finding the main topic -- represented by an article -- of the whole cascade of tweets. In this new task, we report the performance of the different models in dependence on the cascade size.

\end{abstract}

\maketitle

\section{Introduction}
X (formerly Twitter) has emerged as a contemporary agora in recent years, serving as a platform where individuals express their opinions and viewpoints on current events. According to \citet{WhatIsTwitter}, more than 85\% of the topics discussed on this social media platform are directly linked to ongoing news. However, a challenge arises as only a limited number of tweets establish explicit connections to specific events. Public opinion researchers, institutions engaged in countering disinformation, and other entities responsible for monitoring public discourse are compelled to heavily rely on the active participation of humans in identifying novel keywords and emerging events. Hence, the development of automatic tools that could link tweets to related press articles would allow for indirect identification of disinformation and thus have possible applications in fighting the spread of fake news. Disinformation on X is present in various contexts, e.g., medical, political, and societal, and is present in all regions of the world. However, a new 
wave was observed after the aggression of Russia on Ukraine. This wave is not restricted to the English language but is clearly present in other local languages, e.g., Ukrainian, Polish, or Czech. Many examples of fake news in this context describe non-existing events. In order to study this phenomenon and ultimately provide tools to counter it, we concentrate on studying related news and tweets in two languages: Polish and English. 

However, the significant challenge in linking tweets to real events lies in the fact that tweets differ in nature from press articles. They are concise, with a maximum length of 280 characters, and are generally more informal in tone. Initially, topic models \cite{TopicModels} were employed to address this challenge. However, these models proved inadequate in capturing the differences between tweets and articles, as they treated both domains uniformly. Another approach utilized a graph-based latent variable model based on Named Entity Recognition \cite{LinkingTweetsToNewsGraphNER}, but it fell short by focusing primarily on NER while neglecting a significant portion of semantic information from articles and tweets. Another study \cite{TweetRecommender} focused on a supervised trained classifier, but it faced limitations as the classifier was designed for a fixed set of articles and lacked the ability to build a general representation, essential for analyzing new topics. In contrast, \citet{LinkingTweetsWithWordEmbeddings} proposed the use of two embedding spaces based on GloVe \cite{pennington-etal-2014-glove} that were transformed into a new space, where related tweets and news articles would be positioned in close proximity to each other. In particular, none of these papers applied deep neural networks for this task, thus leveraging cutting-edge natural language processing capabilities of existing transformer networks. In this paper, we commence such a study and successfully demonstrate the capabilities of a novel neural network model designed especially for this task. In order to demonstrate the capabilities of the proposed model, we create new large datasets which allow for reliable comparison of our approach with existing ones.

The first contribution of the paper are new annotated datasets linking tweets and articles. We hope that these datasets will serve as a benchmark and reference for further study of this topic. 
We constructed four datasets:
\begin{samepage}
\begin{itemize}
    \item the \emph{Tweets -- PAP News Dataset} -- comprising tweet-article pairs in Polish related to the war in Ukraine, 
    \item the \emph{Tweets -- BBC News Dataset} -- contains tweet-article pairs in English related to the war in Ukraine,
    \item the \emph{Cascades -- PAP News Dataset} -- in this dataset we expand individual tweets coming from \emph{Tweets -- PAP News Dataset} into related information cascades, i.e., quotes and replies,
    \item the \emph{Cascades -- BBC News Dataset} -- here full information cascades have been downloaded on the topic of the war in Ukraine.
\end{itemize}
\end{samepage}
The \emph{Tweets} datasets served the purpose of training, prototyping, and benchmarking models. The \emph{Cascades} datasets are created for the purpose of testing whether it is possible to more accurately find the related article for the whole information cascades.

Inspired by the success of the CLIP model \cite{CLIP} in computer vision, which is based on a contrastive learning approach and models similarities between images and captions, we propose a similar contrastive learning approach to train a representation space where transformed linked articles and tweets exhibit proximity. As the second contribution of this paper, we introduce our contrastive learning approach with pre-trained BERT \cite{devlin2019bert} named \emph{CATBERT} (Contrastive Articles Tweets BERT). An important aspect of contrastive learning is that it was designed to handle automatically generated data and requires only a limited number of manual annotations -- typically only for the creation of the test set. Indeed, we apply it to a dataset where matches are generated through heuristics that are not flawless. Nonetheless, the implementation of contrastive learning, which was demonstrated to be robust in various contexts \cite{RobustContrastiveLearning}, applies as well to our case and proves to be a valuable tool in mitigating the limitations associated with manual data curation. 

In our experiments, we are comparing the \emph{CATBERT} model with 3 other solutions, which generate vector representation of texts, i.e., Latent Dirichlet Allocation (LDA), Term Frequency-Inverse Document Frequency (TF-IDF), and OpenAI embeddings generated via API calls. After computing such vector representations for both articles and tweets, the best matching pairs are computed using cosine similarity. First, we demonstrate that \emph{CATBERT} performed significantly better than other methods on the \emph{Tweets -- PAP News Dataset}. This clearly demonstrates the strength of contrastive learning in this novel context. Second, we study the cascade datasets with the aim to show new approaches for aggregating information over a set of related tweets. With this respect, we concentrate on the study of the English language dataset, which due to its more diverse nature is more challenging. We observe that all developed methods perform relatively weakly. Where the best results are obtained using OpenAI embeddings aggregated appropriately with the max function. Furthermore, our findings demonstrate that aggregating indeed increases the average precision of the methods and it is easier to find related articles for the whole information cascades than for single tweets.
\section{Related work}
\subsection{Different Sources of News Information}
The linkage between tweets and news articles represents only one example of analyzing two distinct mediums of information. In this context, \citet{LinkingTweetsToNewsGraphNER} focuses on leveraging correlations between textual elements, employing features such as hashtags, named entities, and temporal information to establish connections. They further propose a graph-based latent variable model to facilitate this linkage. In a similar vein, \citet{Suarez_2018} incorporates Named Entity Recognition and article text summarization techniques in their methodology for linking tweets and news articles. Expanding upon this approach, \citet{UnifiedRepresentationUsingGraphAndEntities} augments the use of Named Entity Recognition with a graph-based representation.

Furthermore, several studies have considered the inclusion of diverse news providers to detect and track emerging events. For instance, \citet{MELE2019969} extends this line of research to encompass topic detection and tracking. In a different vein, \citet{spitz_exploring_2018} utilizes named entities as a means of aggregating news from multiple streams. Moreover, they employ a graph-based framework to represent the entirety of the content, enabling further downstream analysis.

However, all of the above studies suffer from small test sample sizes, which limits the possibility of rigorously validating the methods. 

\subsection{Multimodal Representation Learning}
Multimodal models play a pivotal role in mapping inputs from diverse data modalities, such as images and text, into a shared representation space \cite{convirt, CLIP}. This framework of data integration has garnered significant interest and excitement in the research community. Nevertheless, the idea of mapping images and texts into a shared embedding space has been explored in earlier works \cite{5540112}. There have been recent efforts in formulating image and text embeddings as metric learning \cite{NIPS2013_7cce53cf}, and multilabel classification \cite{joulin2015learning}. In the context of our work, these approaches could also be used assuming that one modality is the text of the articles and the other is the text of the tweets.
\section{Datasets}
We constructed four datasets: the \emph{Tweets -- PAP News Dataset}, the \emph{Tweets -- BBC News Dataset}, the \emph{Cascades -- PAP News Dataset}, and the \emph{Cascades -- BBC News Dataset}. The \emph{Tweets -- PAP News Dataset} served the purpose of training, prototyping, and benchmarking models. The \emph{Cascades -- PAP News Dataset} served the purpose of benchmarking cascade aggregation functions. The \emph{Tweets -- BBC News Dataset} was used to train models on English texts. Finally, we evaluated the models' performance on the \emph{Cascades -- BBC News Dataset}.

We share all of those datasets for public use. Due to legal reasons, the tweets are represented by their IDs, and news articles by their URLs -- their content is not included. The datasets are available under \url{https://anonymous.4open.science/r/cascades_news_dataset-2211}.

\subsection{Tweets -- PAP News Dataset}
The Russian invasion of Ukraine in 2022 got a lot of attention on social media. Twitter became a key place to share news, especially stories that hadn't been covered by newspapers or TV yet. Later on, some of these stories were proven wrong and called fake news. Identifying news that has not been reported by traditional media is an intriguing endeavor and holds the potential to build models identifying fake news. This unique situation inspired us to document the event by creating a dataset that can be analyzed and researched further. Consequently, we chose to focus on Polish tweets and news from 2022 related to the war in Ukraine. To achieve our goal, we had to choose the source of news, fetch tweets with Twitter API, and choose a strategy to create training and test datasets.

\subsubsection{News}
We decided to use news scraped from the PAP (Polish Press Agency) website. PAP was chosen as the most reliable and the least biased source of information in Polish. Moreover, these articles are concise and informative. To download news only related to the events in Ukraine, we fetched only those containing specific keywords (like "Ukraine", and "Russia", see \ref{sec:polish-keywords} for the full list) -- resulting in 8,153 articles in the selected timeframe (January 1 -- November 5, 2022).

\subsubsection{Training set}
Inspired by CLIP \citep{CLIP} training process, we wanted to find a natural connection between tweets and articles, similar to images with captions. In the case of tweets - these are tweets with URLs to articles; this way we can be sure those tweets refer to the mentioned article. Unfortunately, the number of tweets with a direct link to the PAP article was scarce (around 1,000). We extended this dataset with all replies and quotes (including replies to quotes) to the tweets with PAP URLs - ultimately 6,049 tweet-article pairs. The vast majority of tweets come from the official PAP account on Twitter \footnote{\href{https://twitter.com/PAPinformacje}{https://twitter.com/PAPinformacje}} or replies to tweets from that account.

\subsubsection{Test and validation sets}
To create gold-standard tweet-article pairs, we presented two sets of news-tweet pairs to our data labeler:
\begin{itemize}
    \item Pairs containing an article and a tweet with a hyperlink to the article, or a tweet that is a quote of, or a reply to, the mentioned tweet.
    \item Pairs created by finding relevant tweets to an article by news-related keywords chosen by the data labeler.
\end{itemize}
Subsequently, data labelers were required to assign one of the following three labels to each pair:
\begin{itemize}
\item \textit{match} - the tweet and the article both pertain to the same event,
\item \textit{no match} - the tweet and the article describe distinct events,
\item \textit{skip} - it is indeterminable whether the tweet relates to any event.
\end{itemize}

The outcome yielded a dataset comprising 335 news articles and 2648 tweets. This dataset was subsequently partitioned equally by news into two subsets: the test and validation datasets. The test dataset encompasses 168 news articles and 1260 tweets with 593 \textit{match} and 668 \textit{no match} labels, whereas the validation dataset contains 167 news articles and 1388 tweets with 594 \textit{match} and 800 \textit{no match} labels. It is important to note that no news article or tweet is present in both datasets.

\subsubsection{Inter-Annotator agreement} \label{sec:annotation-agreement}
To evaluate the efficacy of our annotation process, we enlisted 13 people to assess the relevance of 30 tweet-news pairs drawn from the \emph{Tweets -- PAP News Test Dataset}. The Fleiss' kappa is 0.113, which means slight agreement between voters \cite{Landis_1977}. It suggests that the task is non-trivial. These findings align with similar conclusions drawn by \citet{pasi_data_2018}.

\subsection{Tweets -- BBC News dataset}
In order to see how our findings transfer to other languages, we created an analogous dataset with English tweets and news, both from September 2023. 

We chose BBC as the news source, as it is generally considered to be impartial and informative. It also allowed us to scrape news related to the war in Ukraine easily, as there is a tab containing articles on this topic\footnote{\href{https://www.bbc.com/news/world-60525350}{https://www.bbc.com/news/world-60525350}}. We downloaded 101 news articles in total, in the time range from September 1 to September 26, 2023. In this dataset, we used only 45 articles, for which we were able to find relevant tweets.

We downloaded all tweets, with their quotes and replies, from September 15 to September 26 (we were unable to download older tweets due to recent constraints in the Twitter API \cite{twitter-api-restrictions}) containing a URL to one of the downloaded articles. The total number of downloaded tweets is 5,851; 717 of which contained one of the URLs. Tweets with a link to a downloaded article, together with replies and quotes, are automatically labeled as matching, giving us an English training dataset.

\subsection{Cascades}
In studying Twitter interactions, we use the cascade model. Within this paradigm, we define a cascade as a directed acyclic graph, where each node represents an individual tweet. Edges between these nodes are added if one tweet is a reply or a quote of another tweet. The initial tweet is the edge source, and the reply or quote is the target. This structure not only maps the flow of information but also encapsulates the hierarchical and temporal character of communication on the platform. 

\subsubsection*{Cascades construction}
Cascades datasets are constructed in a recursive manner. Starting from the root we check whether it has children (quotes or replies) and if so we add them to the cascade graph. We repeat this operation on newly added nodes recursively until we reach the leaves of the graph. In this manner, the created cascade is a tree.

\subsection{Cascades -- PAP News Dataset}
Based on the PAP test and validation datasets, we created new datasets, which contained partial cascades with root tweets from the original dataset. Partial, because constructed using previously gathered tweets, as we were unable to download these cascades due to changes in Twitter API policies \cite{twitter-api-restrictions}. In the process, all tweets from a given cascade have the same corresponding article as the root tweet. The goal of this dataset is to find the best cascade aggregation function, taking advantage of well-labeled data.

\subsection{Cascades -- BBC News Dataset}
Due to our interest in cascades of varying sizes, we decided to download whole cascades on the topic of the war in Ukraine, focusing on those bigger than a few tweets. For this purpose, we found all tweets from around a week (from September 20 to September 28, 2023) containing keywords concerning the war (see \ref{sec:english-keywords} for the full list), which were recognized as the most relevant by Twitter API. After removing the tweets with potentially very large cascades (we set the limit to 400 replies and quotes in total), we were left with 458 cascades containing 61,912 tweets in total, with a median of 45.5, a mean of 135, and a maximum of 1,318 tweets.

For the news part, we used all the 101 BBC articles that were already downloaded when creating the \emph{Tweets -- BBC News Dataset}.

\subsubsection*{Data labelling}
Firstly, we limited the dataset to the 100 largest cascades. Then we assigned an appropriate news article to each cascade, if such is present in the collection of those we gathered from BBC. Based on this assignment, we constructed the labeled dataset, marking every cascade-article pair as unrelated, unless the article was assigned to the cascade.

\subsection{Data processing}\label{sec:data-processing}
The datasets were processed suitably to the model used and the dataset language. The data processing pipeline could be composed of:
\begin{itemize}
    \item Cleaning -- applied to both tweets and articles, performs basic text cleaning, including: lowercasing, removing unnecessary characters, such as multiple whitespaces, numerical values, and punctuation, removing URLs, short words, and user tags, and processing emojis;
    \item Lemmatization -- applied to both tweets and articles, lemmatizes the given text; has both Polish and English variants.
\end{itemize}
\section{Models}
In an effort to address the challenge of detecting semantic relationships between tweets and articles, we adopt a methodology that involves encoding information from both sources into vector representations and subsequently comparing them using the cosine similarity measure. Thus, the similarity score for each tweet-article pair is a value between -1 and 1. By obtaining the score for every possible pair, we get the similarity matrix, in which the rows represent tweets and the columns represent articles. To produce the classification matrix (with binary values), we evaluate whether the values in the similarity matrix exceed or fall below the similarity threshold.  

    We investigate two distinct approaches to accomplish the task under investigation. The first approach employs a single embedding model, wherein tweets and articles are treated as texts originating from a common domain. As a result, we make use of a single encoder model to embed both tweets and articles into their respective vector representations.
    
    In contrast, the second approach utilizes a dual embedding model -- \emph{CATBERT}. This model employs two distinct encoder models to separately embed articles and tweets, acknowledging the inherent differences in their nature and characteristics. By using separate encoders, we aim to more effectively capture the unique features and nuances of each text type.

To evaluate the performance of the first approach, we employ encoding methods such as Latent Dirichlet Allocation (LDA), Term Frequency-Inverse Document Frequency (TF-IDF), and OpenAI embeddings. They serve as a basis for comparison with the second approach.

Lastly, we introduce models that leverage the aforementioned encoding methods, but with a different focus: instead of working with tweet-article pairs, these models operate on cascades datasets. These models utilize an aggregation strategy.

\subsection{LDA based model}
In this approach, we use Latent Dirichlet Allocation (LDA), a generative probabilistic model for collections of discrete data such as text corpora. After transforming input text with the LDA model, the output might be interpreted as a vector of likelihoods that the given input is related to the corresponding topics. In this way, we generate embeddings for texts from both tweets and news articles.

To achieve the best possible results, we employ data processing involving cleaning and lemmatization (see \ref{sec:data-processing} for the details).

\subsection{TF-IDF based method}
In the center of this approach, there is a TF-IDF vectorizer. TF stands for term frequency, whereas IDF for inverse document frequency. TF-IDF is a method used in information retrieval for assigning each word its importance in a document in the context of whole corpora.

This method involves an extra step in our data preparation process besides its cleaning and lemmatization. Instead of fitting the vectorizer to the whole, possibly lengthy article, we aimed to represent it in a summary. After reviewing the structure of PAP articles, we observed that its first paragraph - always written in bold by journalists - could quite adequately serve as a summary, by containing the key information from the article.

Such prepared summaries of articles, jointly with the tweets' texts, form the corpora for the TF-IDF vectorizer, which is trained on them collectively. The fitted vectorizer is used to transform documents (tweets and articles' summaries) into vectors of weights of words found in a given document. The last step is a pairwise comparison of tweets and summaries using the cosine similarity of beforehand prepared embeddings.

\subsection{OpenAI embeddings}
In this approach, we gather the tweet and news embeddings through the OpenAI API calls. Specifically, we use the \texttt{embeddings} endpoint with the \texttt{text-embedding-ada-002} model \cite{openai-embeddings}. This is a very powerful GPT3-based model trained for diverse NLP tasks on large datasets. It outperforms most of the models listed in the Massive Text Embedding Benchmark (MTEB) Leaderboard \cite{mteb}. 

\subsection{CATBERT}\label{sec:catbert}
In our approach, we create a model that consists of two embedding networks, where each of them computes embedding for different text types (tweets -- short texts, articles) into joint representation space. This way, similarities between tweets and articles can be measured by calculating the cosine similarity of their representation. For the embeddings of both tweets and articles, we utilize pre-trained BERT-based models tailored to the language of the task. Specifically, for English, we employ Roberta \cite{liu2019roberta} for articles and Twitter Roberta Base pre-trained on Twitter data \cite{barbieri-etal-2020-tweeteval} for tweets. For Polish, we use HerBERT \cite{mroczkowski-etal-2021-herbert} for articles and TrelBERT \cite{szmyd-etal-2023-trelbert} for tweets. Finally, we utilize the CLS token to generate embeddings.

Similarly to CLIP \cite{CLIP}, we employ a contrastive learning approach to train the model. Given pairs of matching and non-matching tweet-article combinations, we aim to bring together the representations of matching pairs while pushing apart the representations of non-matching pairs. As a learning objective cosine embedding loss was used, which is computed given input embedding tensors $x_1,\, x_2$ and label tensor $y$ that states whether the inputs are \emph{similar} or \emph{dissimilar} as follows (here \textit{cos} is the cosine similarity function of two embeddings):
\[
    \text{loss}(x, y) = \begin{cases}
        1 - \cos(x_1, x_2) & \text{if $y=1$} \\
        \max(0,\cos(x_1,x_2)) & \text{if $y=-1$}
    \end{cases}
\]

The \emph{Tweets -- PAP News Training Dataset} has only positive tweet-article pairs. Therefore, to generate training pairs with negative examples, we sample articles from the dataset that are unrelated to the current tweet for each tweet. The ratio of positive-to-negative examples serves as a hyperparameter in our approach, and it is independent of the batch size (in contrast to CLIP \cite{CLIP}).

A challenge we faced was handling long texts, as BERT has a context window limited to 512 tokens, while tweets have their constraints and articles do not. To address this issue, we explored several methods and evaluated their performance in different experimental configurations.

\subsubsection{BERT with truncation}
The most straightforward approach involves truncating texts that exceed the 512-token limit.

\subsubsection{BERT with mean embeddings (BELT-like)}
We split tokens of the whole text into chunks. For each chunk, we add special tokens at the beginning and the end and ensure uniform chunk lengths. We stack the chunks and obtain one embedding for each chunk. The final text embedding is obtained by taking the mean of all the chunks' embeddings.  This approach is inspired by BELT \cite{BELT}.

\subsubsection{BERT with data augmentation}
We implement a data augmentation strategy, which involves creating additional examples by splitting articles into parts. Specifically, we take the first 256 tokens of an article as a \textit{header}, which conveys the main idea, and divide the remaining text into 256-token parts.

\subsubsection{Longformer}
We explored the use of the Polish Longformer \cite{beltagy2020longformer, polish-longformer} model, which is designed to handle longer text sequences and has a context window limited to 4096. This surpasses BERT's context window by a factor of 8 and thus can easily fit the whole news article text.

\section{Experiments}

\subsection{Evaluation metrics}
To measure model performance, we use standard metrics, such as average precision, accuracy, precision, recall, and F1-score, using their standard sklearn implementations \cite{scikit-learn}. The first one is score-based, therefore the similarity matrix returned by the model is used, while the rest use the model's classification matrix. In each case, the model's matrix is flattened, as well as the ground truth matrix. 

The ground truth matrix in each cell contains one of three possible values:
\begin{samepage}
\begin{itemize}
    \item 1 -- the corresponding tweet and news article match;
    \item -1 -- the corresponding tweet and news article don't match;
    \item 0 -- it is not known whether the corresponding tweet and news article match.
\end{itemize}
\end{samepage}
Before calculating metrics, the flattened matrices have removed cells corresponding to the 0 values of the ground truth matrix.

\subsection{Human annotators' performance}
To have a good starting point, we check how well human annotations (described in \ref{sec:annotation-agreement}) perform when evaluated on a subset of the \emph{Tweets -- PAP News Test Dataset}. We define the similarity score for the tweet-article pairs as an average of assessment values (assuming the \textit{match} label corresponds to 1, and \textit{no match} to 0). The threshold for the pair matching was 0.5.

Based on this, we obtain the performance metrics presented in Table~\ref{tab:student-results}.

\begin{table}[h]
    \centering
    \begin{tabular}{|l|c|}
    \hline
    Metric & Value \\
    \hline
    Average precision & 0.996 \\
    Accuracy & 0.900\\
    Precision & 1.000\\
    Recall & 0.889 \\
    F1-score & 0.941 \\
    \hline
    \end{tabular}
    \caption{Metrics of human annotators' performance.}
    \label{tab:student-results}
\end{table}

\subsection{Optimization of model parameters}
For each model tested in this paper, we perform hyperparameter optimization. For this purpose, we train the model on the \emph{Tweets -- PAP News Training Dataset}, then evaluate it on the \emph{Tweets -- PAP News Validation Dataset}. Our aim is to maximize the average precision. The optimization process was aided by Optuna~\cite{optuna_2019}, a hyperparameter optimization framework which automates the task. 

In the next step, the similarity threshold is chosen. We iterate over possible values until we find the one maximizing the F1-score of the model on the \emph{Tweets -- PAP News Validation Dataset}.

Various training configurations of \emph{CATBERT} were evaluated. The results of these evaluations are presented in Table \ref{table:performance}. In subsequent sections of the paper, when referring to \emph{CATBERT}, we will specifically be referring to the model trained with a truncation configuration.

\begin{table*}[h]
    \centering
    \begin{tabular}{|l|c|c|c|c|c|}
    \hline
    Configuration name & Average precision & Accuracy & Precision & Recall & F1-score\\
    \hline
    BERT with truncation & \textbf{0.715} & \textbf{0.680} & \textbf{0.638} & 0.740 & \textbf{0.685} \\
    BERT with mean embeddings & 0.681 & 0.665 & 0.618 & 0.756 & 0.680 \\
    BERT with data augmentation & 0.661 & 0.628 & 0.577 & \textbf{0.784} & 0.665 \\
    Longformer & 0.547 & 0.530 & 0 & 0 & 0 \\
    \hline
    \end{tabular}
    \caption{Evaluation of different \emph{CATBERT} configurations.}
    \label{table:performance}
\end{table*}

Final model results on the \emph{Tweets -- PAP News Test Dataset} can be seen in Table \ref{table:test_dataset_results}.

\begin{table*}[h]
\begin{tabular}{|l|l|l|l|l|l|}
\hline
Model             & Average precision & Accuracy       & Precision      & Recall  & F1-score 
\\ 
\hline
TF-IDF            & 0.639             & 0.596          & 0.576          & 0.530          & 0.552  
\\ 
LDA               & 0.589             & 0.578          & 0.541          & 0.685          & 0.604   
\\ 
CATBERT           & 0.715             & \textbf{0.680} & \textbf{0.638} &\textbf{0.740} & \textbf{0.685}    
\\ 
OpenAI Embeddings & \textbf{0.738}    & 0.650          & 0.618          & 0.671          & 0.643     
\\ \hline
\end{tabular}
\caption{Results on the \emph{Tweets -- PAP News Test Dataset} for each model.}
\label{table:test_dataset_results}
\end{table*}

\subsection{Linking cascades to articles}
In order to see how cascades relate to news articles, we used models developed to detect a linkage between a single tweet and a news article and took a two-step approach. First, we ran a model on each tweet from a cascade and calculated its similarity value with each article, forming a collection of similarity vectors. Then we aggregated these similarity vectors to produce a single vector of news similarity for the cascade. We conducted experiments with the following three aggregation functions: \textit{mean}, \textit{median}, and \textit{max} (see Table \ref{table:aggregation_functions}). For each model, we chose the optimal aggregation method for further experiments.

\begin{table}[h]
\begin{tabular}{|l|c|c|c|}
\hline
Model & mean & median & max \\
\hline
TF-IDF & 0.559 & 0.557 & \textbf{0.567}\\
LDA & \textbf{0.602}& 0.600 & 0.595 \\
CATBERT & \textbf{0.624} & 0.622 & 0.620 \\
\hline
\end{tabular}
\caption{Average precision on the \emph{Cascades -- PAP News Validation Dataset} for different models and aggregation functions.}
\label{table:aggregation_functions}
\end{table}

\subsection{Cascades size impact on predictions}
We conducted an experiment whose goal was to check how the size of cascades affects the quality of model predictions. For this purpose, we used the \emph{Cascades -- BBC News Dataset}.

\subsubsection*{Training models}
To adjust the models to the language switch from Polish to English, we trained the models again, this time on the \emph{Tweets -- BBC News Dataset}. The previously optimized hyperparameters were used in the training process, with the hope that the newly trained models would not perform significantly worse than their counterparts trained on Polish data. In the case of the \emph{CATBERT} model, we replaced the pre-trained BERT-based models as described in \ref{sec:catbert}.

\subsubsection*{Cascade cutting}
Given a cascade of size $N$, the cut cascade of size $n \leq N$ is the original cascade with only the $n$ oldest tweets. This approach ensures that each cut cascade represents a real cascade at a certain point in time.

Using cascade cutting, we compare model results on cascades with the same starting tweet and differing sizes. It allows us to accurately measure the impact of the number of tweets in the cascade on the quality of model predictions.

\subsection{Results}
\emph{CATBERT} outperforms all the other models in terms of accuracy, precision, recall, and F1-score on the \emph{Tweets -- PAP News Test Dataset}. The average precision for OpenAI embeddings is slightly higher.

The basic approach of text truncation to train the \emph{CATBERT} model, closely followed by the BELT-inspired, emerged as the most effective method in terms of both metrics' results and training time. Employing the large Longformer model showed the least success in terms of both aspects.

Unfortunately, switching the language from Polish to English was an unsuccessful attempt with all the metrics dropping significantly. The OpenAI model was expected to perform similarly since it is a versatile multi-language model, so this may be caused by the strict approach to labeling the \emph{Cascades -- BBC News Dataset} and thus, the difficulty of this dataset.
However, LDA, TF-IDF, and \emph{CATBERT} models performed markedly worse, while achieving comparable results on the \emph{Tweets -- PAP News Test Dataset}.

OpenAI embeddings with \textit{max} aggregation is the only model that benefits from analyzing larger cascades. It achieves the best performance on cascades of size of around 150 tweets. There is a slight drop in average precision for cascades of size from 10 to 70 though. Other models visibly yield the best results for cascades of size 1 (single tweets). The root tweets are usually the most informative and it can be inferred that the additional noise adversely impacts the performance.

\begin{figure}
    \centering
    \includegraphics[width=0.49\textwidth]{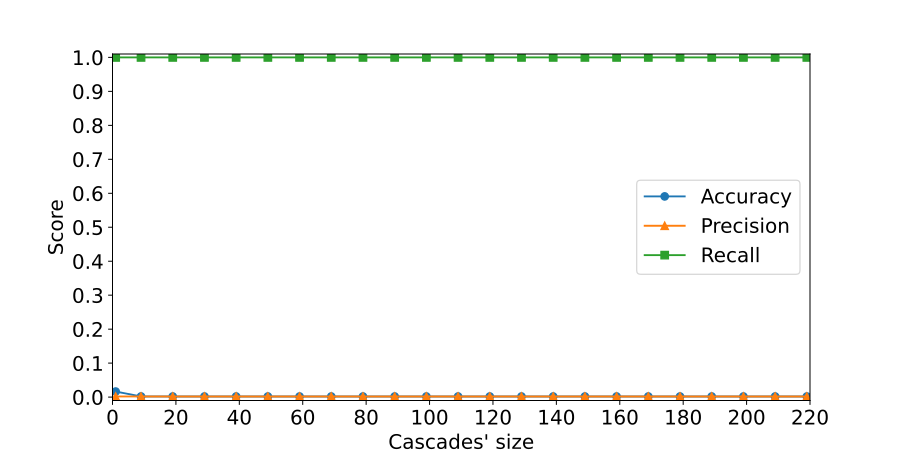}
    \caption{Results on the \emph{Cascades -- BBC News Dataset} for LDA with \textit{mean} aggregation.}
    \label{fig:plot_lda}
\end{figure}

\begin{figure}
    \centering
    \includegraphics[width=0.49\textwidth]{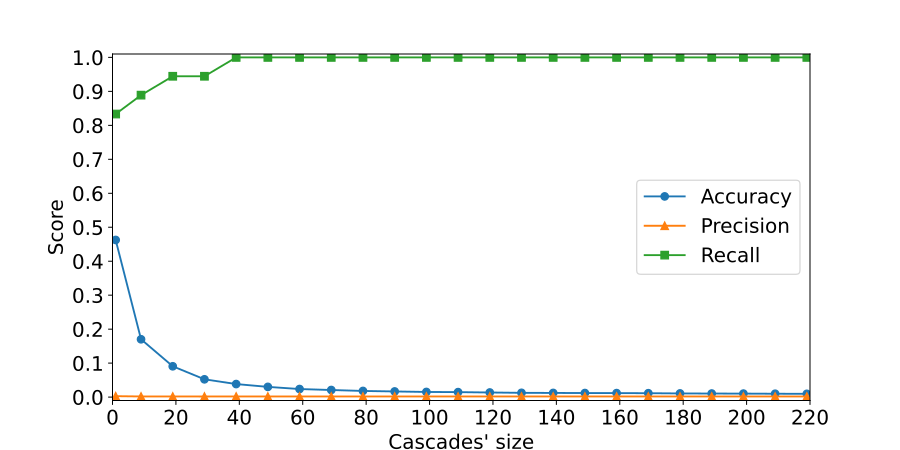}
    \caption{Results on the \emph{Cascades -- BBC News Dataset} for TF-IDF with \textit{max} aggregation.}
    \label{fig:plot_tfidf}
\end{figure}

\begin{figure}
    \centering
    \includegraphics[width=0.49\textwidth]{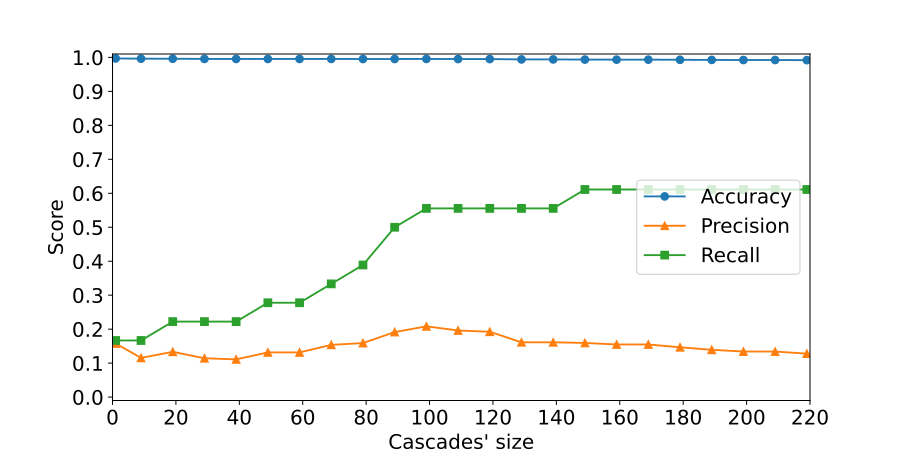}
    \caption{Results on the \emph{Cascades -- BBC News Dataset} for OpenAI embeddings with \textit{max} aggregation.}
    \label{fig:plot_openai_max}
\end{figure}

\begin{figure}
    \centering
    \includegraphics[width=0.49\textwidth]{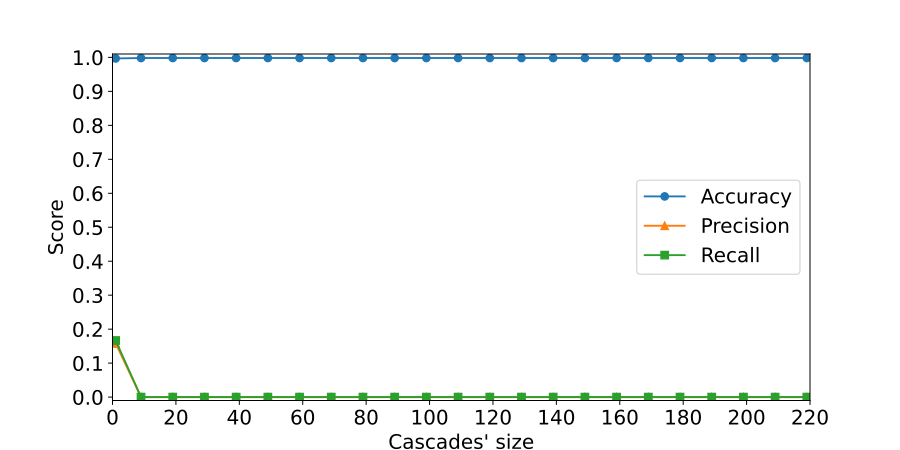}
    \caption{Results on the \emph{Cascades -- BBC News Dataset} for OpenAI embeddings with \textit{mean} aggregation.}
    \label{fig:plot_openai_mean}
\end{figure}

\begin{figure}
    \centering
    \includegraphics[width=0.49\textwidth]{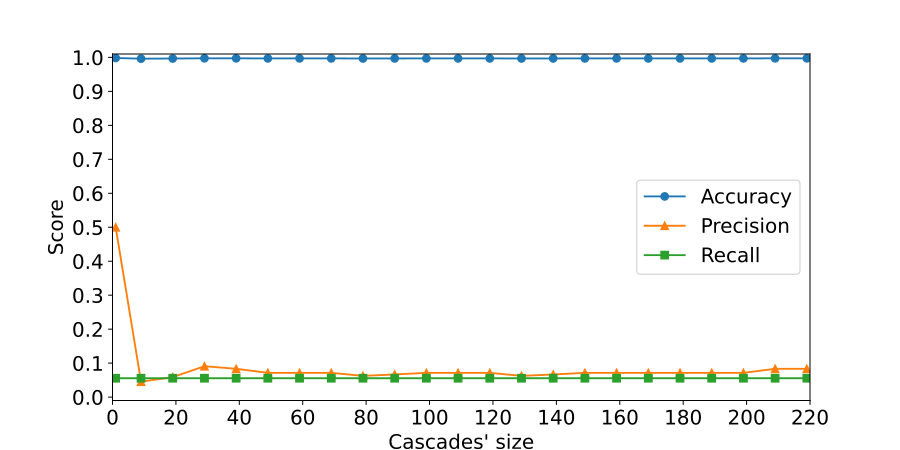}
    \caption{Results on the \emph{Cascades -- BBC News Dataset} for \emph{CATBERT} with \textit{mean} aggregation.}
    \label{fig:plot_catbert}
\end{figure}

\begin{figure}
    \centering
    \includegraphics[width=0.49\textwidth]{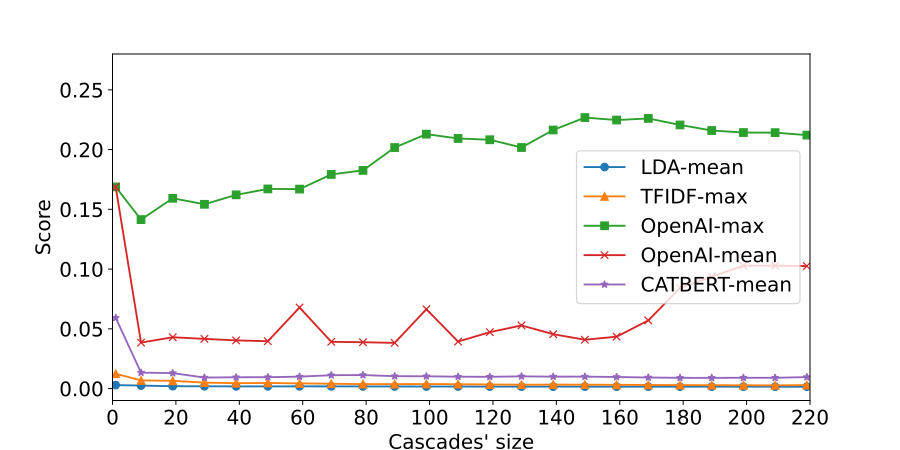}
    \caption{Average precision results on the \emph{Cascades -- BBC News Dataset}.}
    \label{fig:plot_ap}
\end{figure}
\section{Discussion}

In this paper, we introduce an approach, named \emph{CATBERT}, for associating news articles with relevant tweets. Our method leverages contrastive learning to bridge the gap between these two sources of information. To facilitate a comprehensive evaluation of our proposed approach, we have curated datasets in both Polish and English languages. To the best of our knowledge, this is the first attempt to utilize neural networks for this task.

We study different variants of \emph{CATBERT} on both Polish and English language datasets. Furthermore, we showcase the effectiveness of our models when applied to identifying the main topic represented by an article within a cascade of tweets. 

The dataset accompanying this paper is notably larger than those previously available, providing a resource for more rigorous benchmarking. The availability of a larger dataset will enable the development and evaluation of more robust solutions.

Additionally, the challenge we have addressed in this paper can be further explored from the perspective of the \textit{heterogeneity gap}. The dataset utilized in this study poses a challenge for the methods we have evaluated. However, the creation of such a dataset provides the opportunity to benchmark approaches for reducing the heterogeneity gap (as exemplified by \cite{liang_mind_2022}), between mediums not previously considered in such benchmarks.

\section{Acknowledgements}
This work was supported by the National Science Center (NCN) grant no. 2020/37/B/ST6/04179 and the ERC CoG grant TUgbOAT no 772346.

\newpage

\bibliographystyle{ACM-Reference-Format}
\bibliography{main}

\appendix

\section{Keywords used for tweet search}

\subsection{Polish keywords} \label{sec:polish-keywords}

ukraina, ukraiński, rosja, rosyjski, putin, sowiecki, kreml/kremlowski, mińsk, NATO, kijów, moskwa, zełeński, sankcje, rubel, donbas, UKR, RUS, \#ukraine, \#ukraina, \#russia, \#rosja, \#war, \#wojna, \#warinukraine, \#wojnawukrainie, \#wojnanaukrainie, \#standwithukraine, \#ukrainerussiawar, \#putin, \#putin, \#ukrainewar, \#putinwarcrimes, \#ukraineunderattack, \#russianaggression

\subsection{English keywords} \label{sec:english-keywords}

ukraine, ukrainian, russia, russian, putin, soviet, kremlin, minsk, NATO, kyiv/kiev, moscow, zelensky, sanctions, ruble, donbas, UKR, RUS, \#ukraine, \#russia, \#war, \#warinukraine, \#standwithukraine, \#ukrainerussiawar, \#putin, \#ukrainewar, \#putinwarcrimes, \#ukraineunderattack, \#russianaggression

\end{document}